\documentclass{article}

\usepackage{microtype}
\usepackage{graphicx}
\usepackage{caption}
\usepackage{subcaption}
\usepackage{tabularx}
\usepackage{booktabs} 

\PassOptionsToPackage{hyphens}{url}\usepackage{hyperref}

\usepackage[accepted]{mlsys2021}

\usepackage{amsmath}

\newcommand{\RED}[1]{\textcolor{red}{#1}}

\usepackage{lipsum}

\mlsystitlerunning{MicroNets}

\begin{document}

\twocolumn[
\mlsystitle{MicroNets: Neural Network Architectures for Deploying TinyML Applications on Commodity Microcontrollers}



\mlsyssetsymbol{equal}{*}

\begin{mlsysauthorlist}
\mlsysauthor{Colby Banbury*}{arm,harvard}
\mlsysauthor{Chuteng Zhou*}{arm}
\mlsysauthor{Igor Fedorov*}{arm}
\mlsysauthor{Ramon Matas Navarro}{arm}
\mlsysauthor{Urmish Thakker}{sn}
\mlsysauthor{Dibakar Gope}{arm}
\mlsysauthor{Vijay Janapa Reddi}{harvard}
\mlsysauthor{Matthew Mattina}{arm}
\mlsysauthor{Paul N. Whatmough}{arm}

\end{mlsysauthorlist}

\mlsysaffiliation{arm}{Arm ML Research}
\mlsysaffiliation{sn}{SambaNova Systems}
\mlsysaffiliation{harvard}{Harvard University}

\mlsyscorrespondingauthor{Colby Banbury}{cbanbury@g.harvard.edu}

\mlsyskeywords{Machine Learning, MLSys}

\vskip 0.3in

\begin{abstract}
Executing machine learning workloads locally on resource constrained microcontrollers (MCUs) promises to drastically expand the application space of IoT.
However, so-called TinyML presents severe technical challenges, as deep neural network inference demands a large compute and memory budget.
To address this challenge, neural architecture search (NAS) promises to help design accurate ML models that meet the tight MCU memory, latency, and energy constraints. 
A key component of NAS algorithms is their latency/energy model, i.e., the mapping from a given neural network architecture to its inference latency/energy on an MCU. 
In this paper, we observe an intriguing property of NAS search spaces for MCU model design: on average, model latency varies linearly with model operation (op) count under a uniform prior over models in the search space. 
Exploiting this insight, we employ differentiable NAS (DNAS) to search for models with low memory usage and low op count, where op count is treated as a viable proxy to latency. 
Experimental results validate our methodology, yielding our MicroNet models, which we deploy on MCUs using Tensorflow Lite Micro, a standard open-source neural network (NN) inference runtime widely used in the TinyML community. 
MicroNets demonstrate state-of-the-art results for all three TinyMLperf industry-standard benchmark tasks: visual wake words, audio keyword spotting, and anomaly detection. Models and training scripts can be found at \url{https://github.com/ARM-software/ML-zoo}.

\end{abstract}
]



\printAffiliationsAndNotice{\mlsysEqualContribution} 

\section{Introduction}
\label{sec:intro}

Machine learning (ML) methods play an increasingly central role in a myriad of internet-of-things (IoT) applications.
Using ML, we can interpret the wealth of sensor data that IoT devices generate.
Prototypical uses in IoT include tasks such as monitoring environmental conditions such as temperature and atmosphere (e.g., carbon monoxide levels), monitoring mechanical vibrations from machinery to predict failure, or visual tasks such as detecting people or animals.
User interfaces based on speech recognition and synthesis are also very common, as many IoT devices have limited user input features and small displays.
In mobile applications, ML inference is often offloaded to the cloud, where compute resources are more abundant.
However, offloading introduces overheads in terms of latency, energy and privacy.
It also requires access to communications, such as WiFi or cellular access.
For the proliferating class of IoT devices, offloading can be prohibitively expensive, in terms of both the radio chips which increase the bill of materials, as well as the network access costs.

\begin{table}[t]
\caption{
Illustrative comparison of hardware for CloudML, MobileML and TinyML, including the MCUs targeted in this work.
}
\centering
\scriptsize
\begin{tabular}{l| c | c | c | c | c}
\toprule
Platform                   &  Architecture           &   Memory          &  Storage         & Power         & Price \\
\midrule                        
\textbf{CloudML}           & GPU                     & HBM               & SSD/Disk         &               &           \\
Nvidia V100                & Nvidia Volta            & 16GB              & TB$\sim$PB       & 250W          & \$9K    \\
\midrule                    
\textbf{MobileML}          & CPU                     & DRAM              & Flash            &               &           \\
Cell Phone                  & Mobile CPU             & 4GB               & 64GB             & $\sim$8W      & $\sim$\$750    \\
\midrule                    
\textbf{TinyML}            & MCU                     & SRAM              & eFlash           &               &        \\
\citealt{STM32F446RE}       &  Arm \citealt{arm-m4}   & 128KB             &  0.5MB           & 0.1W   & \$3    \\
\citealt{STM32F746ZG}       &  Arm \citealt{arm-m7}   & 320KB             &  1MB             & 0.3W   & \$5    \\
\citealt{STM32F767ZI}       &  Arm \citealt{arm-m7}   & 512KB             &  2MB             & 0.3W   & \$8    \\
\bottomrule
\end{tabular}
\vspace{-15pt}
\label{table:mcus}
\end{table}

\begin{figure}[t]
\centering
    \includegraphics[width=\linewidth]{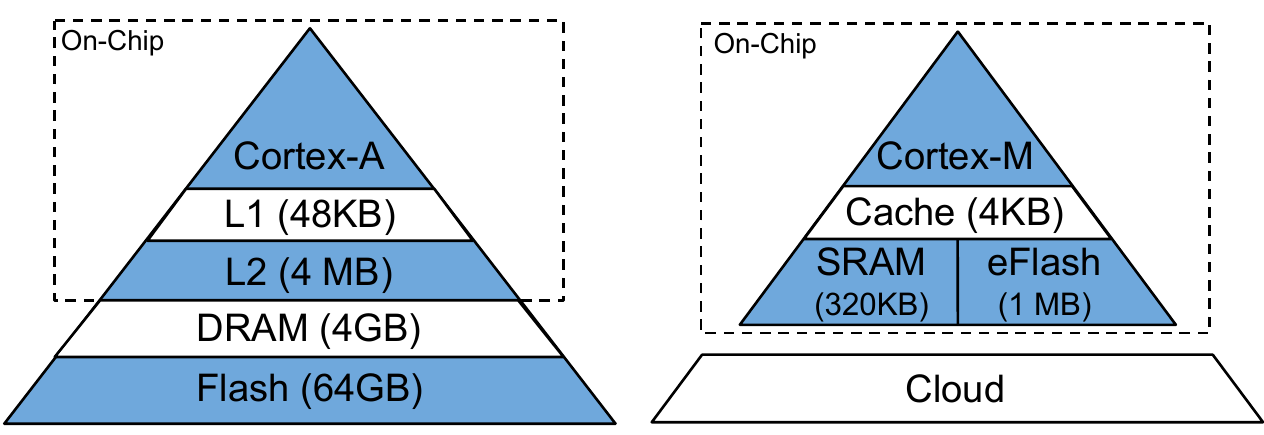}c
    \footnotesize
    (a) Mobile SoC
    \hspace{65pt}
    (b) MCU
     \hspace{12pt}
    \vspace{-6pt}
    \caption{Illustration of memory hierarchies for (a) a mobile SoC which has a deep memory hierarchy with many levels of on-chip cache and a large off-chip DRAM main memory, and (b) an MCU with a flat on-chip memory system with no off-chip main memory.
    }
    \label{fig:memory-hierarchy}
\vspace{-6pt}
\end{figure}

\textit{TinyML} is an alternative paradigm, where we execute ML tasks locally on IoT devices.
This allows for real time analysis and interpretation of data at the point of collection, which translates to huge advantages in terms of cost and privacy.
Microcontroller units (MCUs) are the ideal hardware platform for TinyML, as they are typically small ($\sim$1cm$^3$), cheap ($\sim$\$1) and low-power ($\sim$1mW) compared to mobile and cloud platforms (Table~\ref{table:mcus}).
MCUs typically integrate a CPU, digital and analog peripherals, on-chip embedded flash (eFlash) memory for program storage and Static Random-Access Memory (SRAM) for intermediate data.
However, deploying deep neural networks on MCUs is extremely challenging; the most severe limitation being the small and flat memory system (Figure~\ref{fig:memory-hierarchy}) within which the model weights and activations must be stored.
Therefore, to achieve the promise of TinyML, we must aggressively optimize models to best exploit the limited resources provided by an MCU hardware and software stack.

Mounting interest in TinyML has led to some maturity in both software stacks and benchmarks.
The open source TensorFlow Lite for Microcontrollers (TFLM) inference runtime~\cite{david2020tensorflow} allows for straightforward and portable deployment of NNs.
TFLM uses an interpreter to execute an NN graph, which means the same model graph can be deployed across different hardware platforms.
When compared to code generation based methods~\cite{utensor}, TFLM provides portability across MCU vendors, at the cost of a fairly minimal memory overhead.
Recently, the ML performance (MLPerf) benchmarking organization has outlined a suite of benchmarks for TinyML called TinyMLPerf~\cite{banbury2020benchmarking}, which consists of three TinyML tasks of visual wake words (VWW), audio keyword spotting (KWS), and anomaly detection (AD).
Standardizing TinyML research results around a common open-source runtime and benchmark suite makes comparing research results easier and fairer, hopefully driving research progress.


Previous work on TinyML has largely considered model design without consideration for the real deployment scenario (e.g. SpArSe~\cite{fedorov2019sparse}, Structured Matrices \cite{urmish1,9027218}), or has used closed-source software stacks which make deployment and comparison impossible (e.g. MCUNet~\cite{lin2020mcunet}).
In this paper, we describe MicroNets, a family of models which can be deployed with publicly available TFLM, for the three TinyMLperf tasks of VWW, KWS and AD.
In contrast to previous TinyML work that uses black-box optimizations, such as Bayesian optimization~\cite{fedorov2019sparse}, and evolutionary search~\cite{lin2020mcunet}, MicroNets are optimized for MCU inference performance using differentiable neural architecture search (DNAS).

The contributions of this work are summarized below.
\begin{itemize}

    \item Using an extensive characterization of NN inference performance on three representative MCUs, we demonstrate that the number of operations is a viable proxy for inference latency and energy.
    
    \item We show that differentiable neural architecture search (DNAS) with appropriate constraints can be used to automatically construct models that fit the MCU resources, while maximizing performance and accuracy. 
    
    \item We provide state of the art models for all three TinyML tasks, deployable on standard MCUs using TFLM.
\end{itemize}

\section{Related Work}
\label{sec:related}
Since its inception, deep learning has been synonymous with expensive, power-hungry GPUs~\cite{krizhevsky2012alexnet}.
However, the current interest in deploying deep learning models on MCUs is reflected in a small number of papers that have begun to explore this promising space.
In this section, we briefly survey the literature related to TinyML, divided between hardware, software and machine learning.

\paragraph{Hardware}
The current interest in ML has led to a growing demand for arithmetic compute performance in MCU platforms, which was previously driven by digital signal processing workloads.
Single-instruction multiple-data (SIMD) extensions~\cite{pandh-2011} are one of the most effective approaches to achieving this in the CPU context, but increase silicon area and power consumption.
The Arm Helium extensions~\cite{arm-helium} address this using a lightweight SIMD implementation targeted to MCUs.
Beyond CPUs, various accelerators~\cite{gap8-asap18,kodali-iccd17,whatmough-jssc18,fixy2019sysml} and co-processors such as digital signal processors (DSPs)~\cite{efland2016high} and micro neural processing units (uNPUs)~\cite{arm-u55} typically offer greater performance and energy efficiency, at the cost of a more complex and less portable programming model~\cite{zhu-mlsys19}.
Finally, optimized memory technologies~\cite{mem-li-dac19} and subthreshold circuit techniques 
~\cite{ambiq-spot} can be used to reduce the power consumption at the circuit level.
In this work we specifically target commodity MCUs (Table~\ref{table:mcus}), and expect results to improve with new hardware generations.

\begin{figure}[t]
\centering
    \includegraphics[width=0.8\columnwidth]{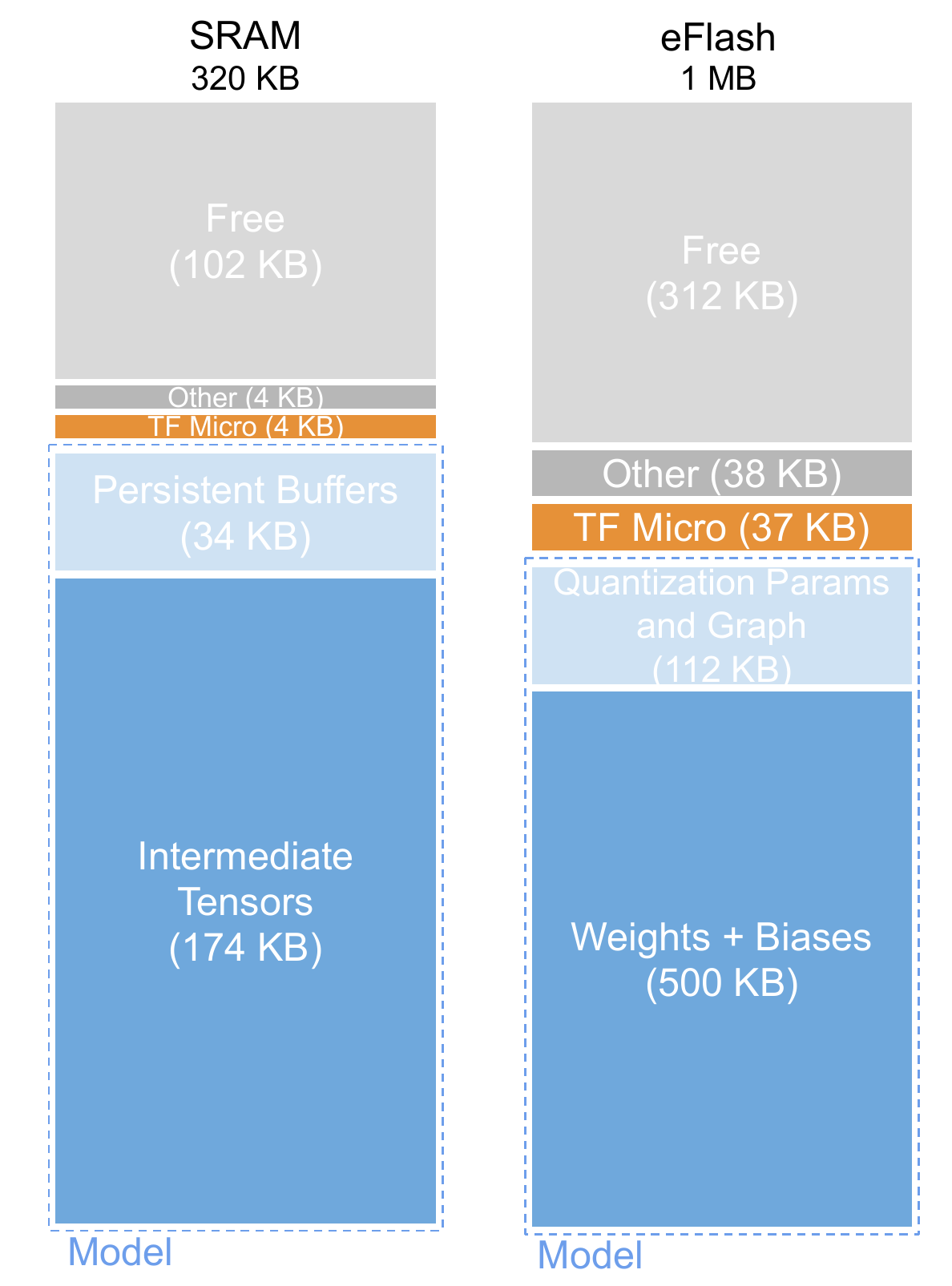}
    \vspace{-8pt}
    \caption{
    Breakdown of SRAM and eFlash memory occupancy for a KWS model on TFLM runtime on the STMF746ZG.}
    \label{fig:mem-map}
\end{figure}

\paragraph{Software}
A critical element for TinyML is the inference software stack. 
MCU vendors typically provide low-level libraries with specific primitives for basic NN operators like convolution, such as the CMSIS-NN~\cite{lai2018cmsis} kernels which are optimized for Arm Cortex-M devices. 
Alternatively,~\citealt{micro-tvm} automatically generates low level kernels. 
These kernels need stitching together to implement a neural network inference graph.
Popular ML frameworks like~\citealt{tensorflow} and~\citealt{pytorch} are unsuitable for inference on MCUs, as the memory requirements are too large.
A number of ML inference runtimes have emerged to fill this need on MCUs.
There are two fundamental approaches: code generation and interpreter. 
The code generation approach takes a model definition and automatically generates C code directly. 
In general, this approach typically gives the best results, but the generated code is not portable between different platforms.
Examples include~\citealt{utensor}, tinyEngine~\cite{lin2020mcunet}, and embedded learning library~\cite{ell}.
In contrast, TensorFlow Lite for Microcontrollers~\citealt{TFLM} is an interpreter based runtime for executing TensorFlow Lite graphs on MCUs. 
TFLM supports most common NN layers, with the notable exception of recurrent networks. 
It is widely supported by hardware vendors and supports many optimized kernels on the back end for specific platforms. 
Compared to code generation based methods, TFLM is more portable but has some overheads. 
We use TFLM due to its portability, ease of deployment and open-source nature.

\paragraph{Machine Learning}
The challenges with implementing CNNs on MCUs were discussed in Bonsai~\cite{kumar2017bonsai}, namely that the feature maps of typical NNs require prohibitively large SRAM buffers. As a more storage efficient alternative to CNNs, pruned decision trees were proposed to suit the smallest MCUs with as little as 2KB of SRAM. \citet{gupta2017protonn} propose a variant of k-nearest neighbors tailored for MCUs. \citet{pmlr-v97-gural19a} propose a novel convolution kernel, reducing activation memory and enabling inference on low-end MCUs. SpArSe~\cite{fedorov2019sparse} demonstrated that by optimizing the model architecture, CNNs in fact can be deployed on MCUs with SRAM down to 2KB.
This was achieved using NAS, which has emerged as a vibrant area of research, whereby ML algorithms construct application specific NNs to meet very specific constraints~\cite{elsken_neural2019}. 
SpArSe employs a Bayesian optimization framework that jointly selects model architecture and optimizations such as pruning. 
Similarly, MCUNet~\cite{lin2020mcunet} uses evolutionary search to design NNs for larger MCUs (2MB eFlash / 512KB SRAM) and larger datasets including visual wakewords (VWW) \cite{chowdhery2019visual} and keyword spotting (KWS) \cite{warden2018speech}). 
Reinforcement learning (RL) has also been used to choose quantization options in order to help fit an ImageNet model onto a larger MCU (2MB eFlash)~\cite{rusci2020leveraging}.
As well as images, audio tasks are an important driver for TinyML. TinyLSTMs~\cite{fedorov2020tinylstms} shows that LSTMs for speech enhancement in smart hearing aids are similarly amenable to deployment on MCUs, after targeted optimization.


In this paper, we use differentiable NAS (DNAS)~\cite{liu2018darts} to design specialized MCU models to target the three TinyMLperf tasks.
Unlike black-box optimization methods that have previously been applied to TinyML problems, like Bayesian optimization~\cite{fedorov2019sparse} and evolutionary search~\cite{lin2020mcunet}, DNAS uses gradient descent and lends itself to straightforward implementation in modern auto-differentiation software like Tensorflow with acceleration on GPUs. 
Our work provides experimental evidence that DNAS is capable of satisfying MCU-specific model constraints, including eFlash, SRAM, and latency. In contrast to \cite{lin2020mcunet}, our work uses a standard deployment framework (TFLM).

\section{Hardware Characterization}
\label{sec:hardware}

\subsection{System Overview}
\label{sec:hardware:system}


In this section we characterize the performance of NN inference workloads on MCUs.
The MCUs we use (Table~\ref{table:mcus}) are fairly self-contained, consisting of an Arm Cortex-M processor, SRAM for working memory, embedded flash for non-volatile program storage, and a variety of digital and analog peripherals.
Unlike their mobile, desktop and datacenter counterparts, MCUs have a rather flat memory system, as illustrated in Figure~\ref{fig:memory-hierarchy}.
Mobile and cloud computer systems universally employ a large off-chip main memory (usually DRAM).
However, MCUs are typically equipped with only on-chip memory, which is relatively small to keep the die size reasonable.
Figure~\ref{fig:mem-map} gives an example memory map showing how a KWS model is mapped onto the STM32F746ZG devices by TFLM.
Activation buffers are allocated in the SRAM, while the model weights and biases and graph definition are allocated in the eFlash memory.
Alternatively, weights can be stored in SRAM, but we found experimentally that this results in only about a 1\% speedup in end-to-end latency, while significantly reducing the space available for activations, which cannot be stored in eFlash.

In terms of throughput, this flat memory system coupled with the lower clock frequencies and simple (cheap) microarchitectures used in MCUs results in a predominately compute-bound system.
The Cortex-M7 can dual issue load and ALU instructions, which the Cortex-M4 cannot.
This gives higher IPC, which, combined with a 20\% higher clock rate, makes the STM32F646ZG and the STM32F767ZI approximately twice as fast as the STM32F446RE.

Note that the runtime overhead for the TFLM interpreter is fairly minimal, requiring just 4KB of SRAM and 37 KB of eFlash. The 34KB SRAM block labeled as persistent buffers in Figure \ref{fig:mem-map} scales with the size of the model and contains buffered quantization parameters and the structs that hold pointers to the intermediate tensors and to the operators. 

\subsection{Layer Latency}

\begin{figure}[t]
    \includegraphics[width=\linewidth]{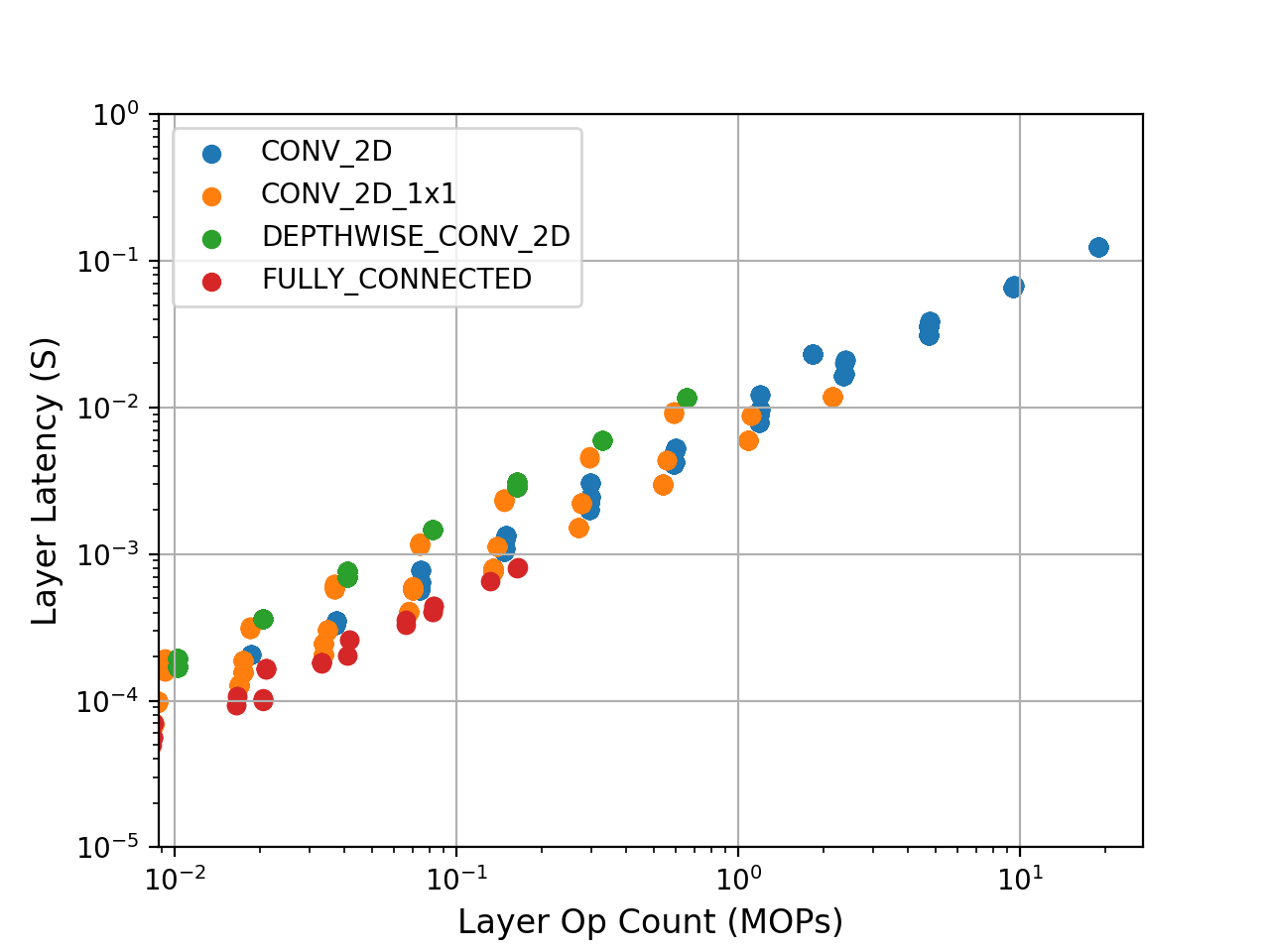}
    \vspace{-20pt} 
    \caption{
    Measured latency of a range of different individual layer types and sizes on the STM32F767ZI using TFLM.
    Different layers can exhibit a spread in latencies for the same ops count, due to variations in, for example, data reuse and IM2COL overheads.
    }
    \label{fig:per-layer}
\end{figure}

In this section, we examine the hardware performance of typical NN layers.
To do this, we generate a large number of layer types\footnote{Excluding RNN layers, not currently supported in TFLM.} and sizes and characterize them on the hardware.
Figure~\ref{fig:per-layer} shows the measured latency of each layer in TFLM and CMSIS-NN kernels, as a function of the number of operations\footnote{A single multiply-accumulate is defined as two operations.} (ops).
We observe that different layer types and sizes result in some spread in throughput, which was previously observed by~\citet{lai2018notallops}.
2D convolutions and fully connected layers exhibit lower latency per op than depth-wise convolutions. 
This is likely due to depth-wise convolutions having less operations relative to their IM2COL overhead.
We also note some variability in ops/s between 2D convolution layers. This is primarily caused by the sensitivity of the CMSIS-NN kernel to input and output channel sizes. 
The CMSIS-NN CONV 2D kernel is substantially faster when the number of input and output channels are divisible by four.
As an example, we observe that \textit{increasing} the input/output channels of a convolution layer from 138/138 to 140/140 \textit{decreases} the latency from 37.5ms to 21.5ms (57\% speedup).

\subsection{Model Latency}

\begin{figure}[t]
    \includegraphics[width=\linewidth]{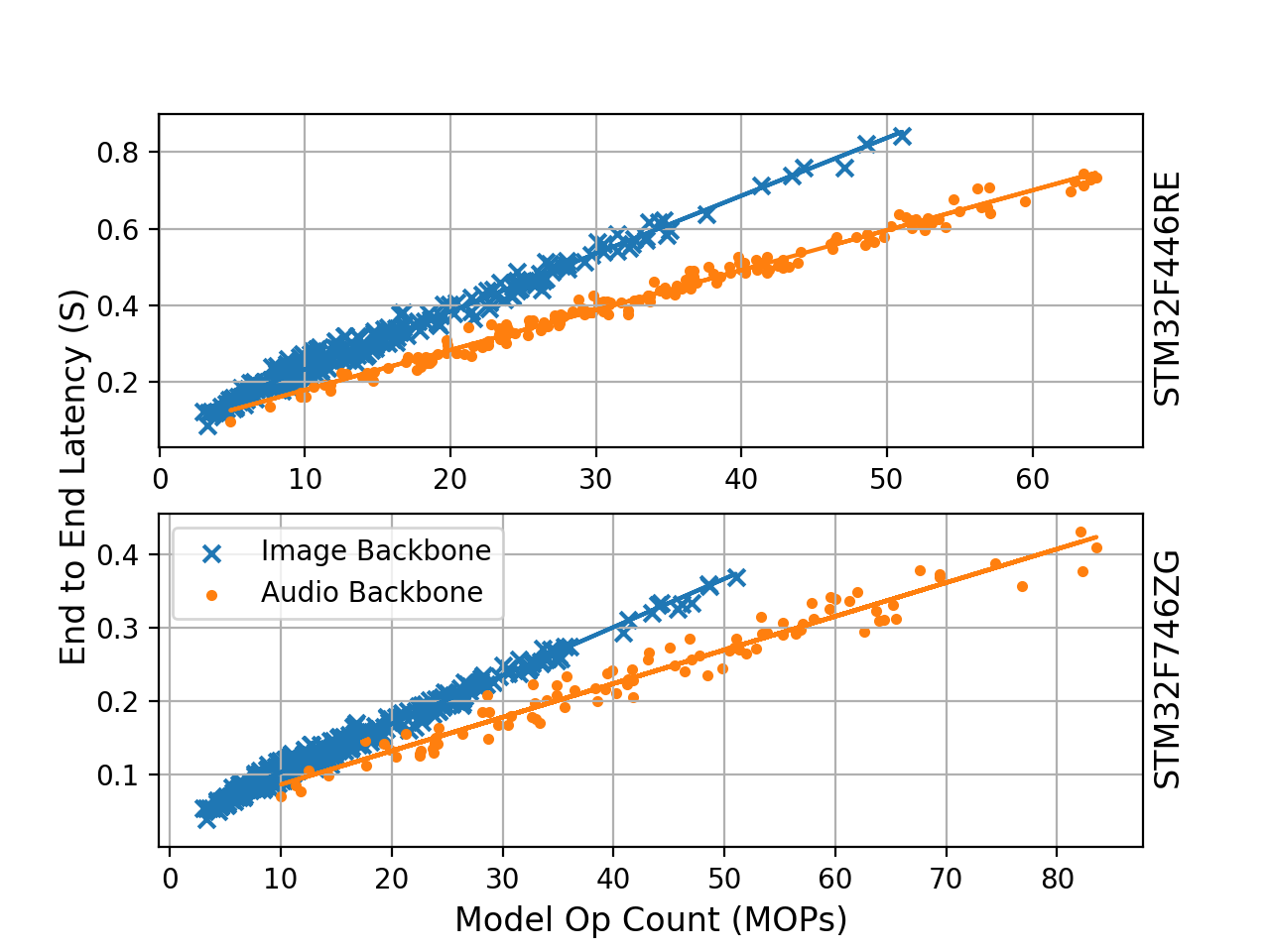}
     \vspace{-20pt}
     \caption{
     Measured latency of whole models randonly sampled from two backbones, on STM32F446RE and STM32F746ZG.
     Models sampled from a given search space exhibit latency linear with ops, despite the variation seen with individual layers.
     }
     \label{fig:model_latency_vs_ops}
\end{figure}

Next, we characterize whole end-to-end models.
To do this, we setup a parameterized supernet backbone that we randomly sample.
This allows us to automatically generate a large number of random models with different layer types and dimensions, which we then characterize on the hardware in terms of latency and, in the next subsection, energy.
Figure~\ref{fig:model_latency_vs_ops} shows measured model latency on the STMF446RE and the STMF746ZG.
Measurements are shown for random models sampled from backbones tailored to two different tasks \textit{viz.} image classification and audio classification.

Interestingly, the measured latency for the end-to-end models is linear with op count ($0.95 < r^2 < 0.99$).
This is perhaps surprising given the variation seen with the layer-wise latency measurements (Figure~\ref{fig:per-layer}).
Also, we observe that models sampled from the two different backbones results in a different slope.
The explanation for this is that although single layers exhibit variation in latency as a function of ops, in a whole model this is averaged across many layers.
Since a given search space will typically be dominated by a particular layer type, in terms of ops, the result is that we see a linear latency \textit{for models sampled from the same backbone}.
The KWS backbone has $\sim$40\% higher throughput (Mops/S) than the CIFAR10 backbone, which is due to the mix in layer types and sizes.
Finally, STM32F746ZG is around twice as fast as STM32F446RE (Section~\ref{sec:hardware:system}).



\subsection{Model Energy}
\begin{figure}[t]
    \includegraphics[width=\linewidth]{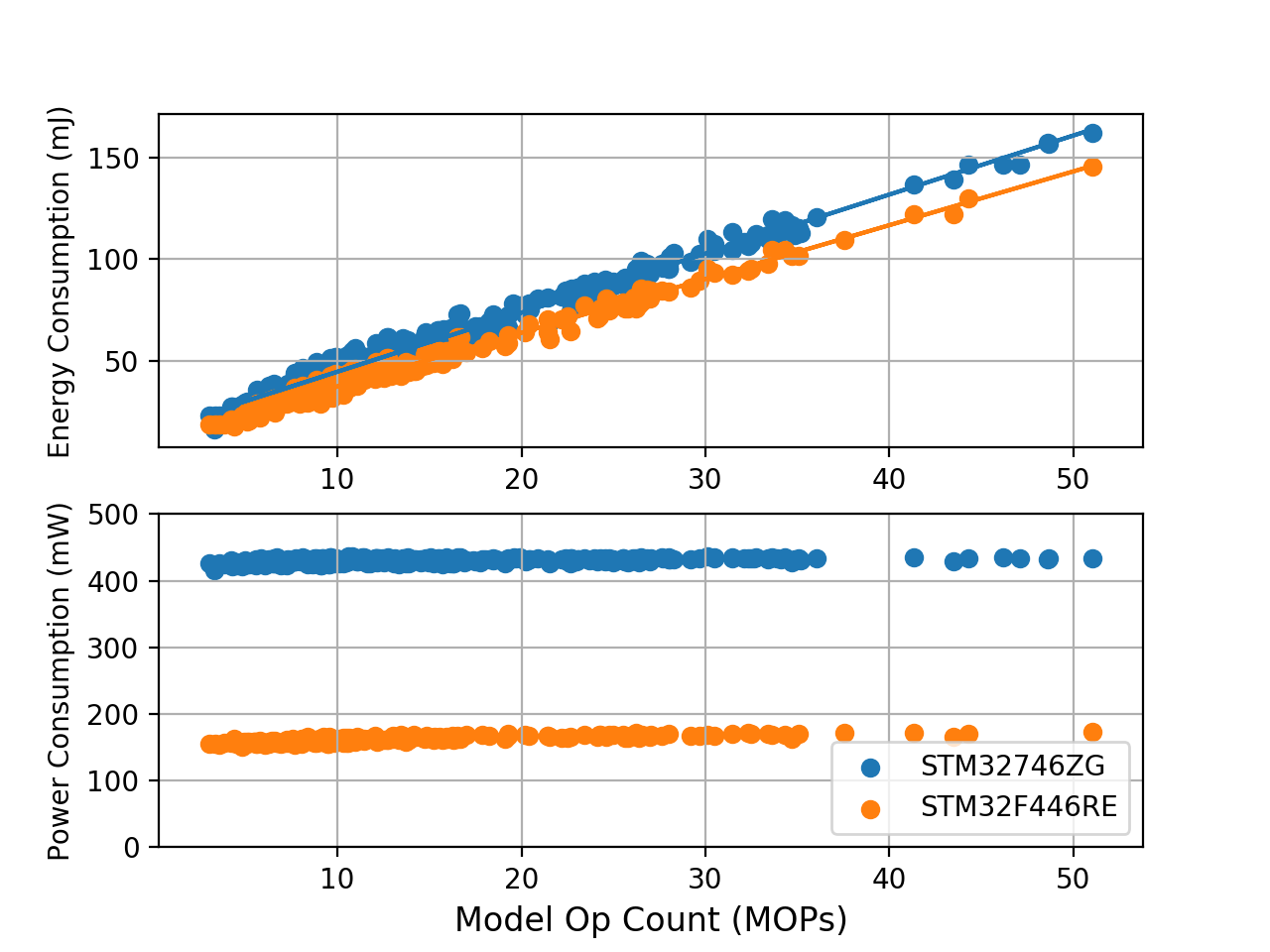}
     \vspace{-20pt}
     \caption{Measured energy and power of models randomly sampled from an image classification CNN backbone.
     MCUs have simple microarchitectures and memory systems and hency power is fairly constant.
     Therefore, energy is largely determined by latency, which is in turn a linear function of model ops.
     }
     \label{fig:model_energy_power}
\end{figure}

Energy consumption is obviously a critical metric for TinyML.
Following the same random model sampling methodology used to characterize latency, we measured the current consumption of 400 models from the CIFAR10 backbone.
We use the Qoitech Otii Arc~\cite{otii} to power the MCU boards and measure the current draw with the inference workload looping.
Figure~\ref{fig:model_energy_power} shows the average power consumption versus the op count of each model on two MCUs.
Clearly, there is little variance in power consumption between models ($\sigma/\mu = 0.00731$), i.e. power is essentially independent of model size or architecture.
Additionally, Figure~\ref{fig:model_energy_power} shows the energy consumption versus the op count of each model. We observe that executing the same model on a smaller MCU reduces the total energy consumption despite an increase in latency. 
This decreased energy consumption motivates the design of models that can fit within the tighter constraints of smaller devices.

\subsection{Summary}

Below is a summary of the findings of this section:
\begin{itemize}
    \item Although ops is not a good predictor for the latency of a single layer, it \textit{is} a viable proxy for the latency of an entire model sampled from a given backbone.
    \item For a given MCU, power is largely independent of model size and design.
    Therefore, energy per inference is a function of the size of the MCU, which determines power, and the number of ops, which dictates latency.
\end{itemize}

Therefore, when designing a model from within a backbone for a given task, ops is a viable proxy for both latency and energy, as measured on the target hardware and software.




\section{TinyMLperf Benchmark Tasks}
\label{sec:benchmarks}

This section describes the TinyMLPerf benchmark tasks: Visual Wake Words (VWW), Keyword Spotting (KWS), and Anomaly Detection (AD).
These were selected by a committee from industry and academia, to represent common TinyML application domains~\cite{banbury2020benchmarking}.

\subsection{Visual Wake Words}
The VWW dataset used in TinyMLperf is a visual classification task, where each image is labeled as $1$ when a person occupies at least 0.5\% of the frame and $0$ when no person is present \cite{chowdhery2019visual}. The dataset contains 82,783 train and 40,504 test images, which we resize to a common resolution of $224 \times 224$. We use the standard ImageNet data prepropecessing pipeline. 

\subsection{Audio Keyword Spotting}
Audio KWS (a.k.a. wake words),
finds application in a plethora of use cases in commercial IoT products (e.g. Google Assistant, Amazon Alexa, etc.). 
Recent research has explored various model architectures suitable for resource constrained devices~\cite{kwsgnn,kwsatt,KusupatiSBKJV18,kron1}.
Among these, CNNs achieve good accuracy \cite{kwscnn1,kwscnn2,zhang2017hello,GopeMLSys2019} and have the advantage of being deployable on commodity hardware using existing software stacks.
The KWS dataset in TinyMLperf is Google Speech Commands (V2) \cite{warden2018speech}.
A model trained on this dataset is required to classify an $1$-second long incoming audio clip from a vocabulary of $35$ words into one of the $12$ classes--$10$ keyword classes along with ``silence'' (i.e. no word spoken), and an ``unknown'' class, which is the remaining $25$ keywords from the dataset. 
The raw time-domain speech signal is converted to 2-D MFCC (Mel-frequency cepstral coefficients).
$40$ MFCC features are then obtained from a speech frame of length $40$ms with a stride of $20$ms, yielding an input dimension of $49\times10\times1$ features for $1$ second of audio.
Training samples are augmented by applying background noise and random timing jitter to provide robustness against noise and alignment errors. We follow the same input data processing procedure described in~\citet{helloedge}
 and~\citet{mo2020neural} for training the baselines and other DNAS variants.

\subsection{Anomaly Detection}
Anomaly detection is a task that classifies temporal signals as ``normal" or ``abnormal". Anomaly detection finds numerous application, including industrial factories, where it is deployed on smart sensors to monitor equipment and detect problems. 
The dataset used for anomaly detection in TinyMLperf is MIMII(Slide Rail) \cite{purohit2019mimii}.
It is a dataset of industrial machine sounds operating under normal or anomalous conditions recorded in real factory environments. The original dataset contains different machine types, but we focus on the \textit{Slide Rail} task as selected in TinyMLPerf benchmarks. 

Anomaly detection is an unsupervised learning problem. The model only sees ``normal" samples at training time and is expected to make predictions on a mix of normal and abnormal cases at test time. Many unsupervised learning methods can be applied, however, inspired by state-of-the-art solutions \cite{Giri2020}, we reformulate the problem as a self-supervised \cite{hendrycks2019using} learning problem, so that it can be handled in a similar way as the other two tasks.
The essential idea is to leverage machine ID metadata provided in this dataset. The training dataset contains 4 different machine IDs, each corresponding to a different slide machine for which the audio is recorded. We train a classifier in a supervised way to identify the machine ID given the audio as input. The classifier needs to learn useful information about the normal operating sound of these machines to tell them apart, which can then be used to detect anomaly. At testing time, we use the softmax score for the test sample machine ID as an index of how confident the classifier is about the test sample falling into the normal operating regime data on which it has been trained. Therefore its negative can be used as an anomaly score (higher meaning more likely to be abnormal). The area under the curve (AUC) metric from the receiver operating characteristic (ROC) is calculated using this anomaly score.

Data preprocessing is done in a similar way as for KWS: the audio signal is transformed into log-Mel spectrograms, which are then input to a CNN classifier. An audio clip of length 10s is split into overlapping frames of length 64ms with a stride (hop length) of 32ms between frames. 64 MFCC features are extracted for each frame. The preprocessed dataset is available on Kaggle~\cite{kaggle-ad}.
We then stack 64 frames together to get 64 by 64 images and the next image has an overlap of 44 frames. We found that CNNs can tolerate even lower resolution spectrograms so the image is further down-sampled to 32$\times$32 using bilinear interpolation. This is the input to our CNN classifiers.

\section{MicroNet Models}
\label{MicroNet Models}

\subsection{Optimizations}
We use DNAS to discover models which are highly accurate, while also satisfying SRAM, eFlash, and latency constraints. In the following, we briefly review DNAS and how it can be applied to ML model design for MCUs. For further information, we refer the reader to \cite{liu2018darts,cai2018proxylessnas,dong2019network,Wan_2020_CVPR}. The search begins with the definition of a supernet consisting of decision nodes. The output of a decision node expresses a choice between $K$ options
\begin{align}
    y = \sum_{k=1}^K z_k f_k(x,\theta_k), \; \; \sum_{k=1}^K z_k = 1
\end{align}
where $x$ is the input tensor, $f_k()$ is the operation executed by choice $k$ and parameterized by $\theta_k$, $K$ is the total number of options for the decision node, and $z_k \in \lbrace 0,1 \rbrace$ represents the selection of one of $K$ options. The goal of the search is to select $\boldsymbol{z} = \begin{bmatrix}
z_1 & \cdots & z_K
\end{bmatrix}$ for all of the decision nodes in the supernet. In the present work, we restrict our search to the width for each layer and the overall depth. In this case, each option $f_k()$ represents an operation with a different number of channels \cite{Wan_2020_CVPR} or the identity.

\subsubsection{Optimizing for MCU Memory}
Without any model constraints, DNAS may produce models which violate one or more MCU hardware limits. Given the model size and the size of the intermediate activations produced by modern NNs, eFlash and SRAM play an important role in model design. We incorporate appropriate regularization terms in our DNAS experiments such that the selected models both fit in eFlash memory and produce activations which can fit in available SRAM. For model size considerations, we express the size of a particular selection from the supernet using
\begin{align}\label{eq:node size}
    \sum_{k=1}^K z_k \left \vert \theta_k \right \vert
\end{align}
where $\vert \theta_k \vert$ denotes the cardinality of $\theta_k$. Summing the size of each node, we obtain the size of the supernet as a function of decision parameters $\boldsymbol{z}$ for each decision node, which we use to regularize the DNAS such that the selected architecture meets the MCU eFlash constraint. 

To ensure that the selected architecture satisfies SRAM constraints, we adopt the working memory model from \citet{fedorov2019sparse}, which states that the working memory required for a particular node with inputs $\lbrace x_1,\cdots, x_N \rbrace$ and outputs $\lbrace y_1, \cdots, y_M \rbrace$ is given by $\sum_{n=1}^N \left \vert x_n \right \vert + \sum_{m=1}^M \left \vert y_m \right\vert$. For tensors which are outputs of decision nodes, we replace $\left \vert x_n \right \vert$ by \eqref{eq:node size}. The total model working memory is then defined as the maximum over the working memory of every network node, which we include in the DNAS objective function such that the discovered architecture meets the MCU SRAM constraint. We define the constraint as the available SRAM minus the expected TFLM overhead.

\subsubsection{Optimizing for Latency}
In addition to making sure that discovered models are deployable, we also incorporated a latency constraint into our DNAS experiments. Due to the (almost) linear relationship between latency and number of operations for ML inference on MCUs, we treat the operation count as a strong proxy for latency during optimization. As with memory, we begin by defining the operation count of each decision node as a function of the decision vector $\boldsymbol{z}: \sum_{k=1}^K z_k c_k$, where $c_k$ is the number of ops required to execute option $k$. Note that the number of operations for a particular option typically depends on the input and output tensor sizes, which are a function of decision parameters $\boldsymbol{z}$ \cite{Wan_2020_CVPR}.

\subsubsection{Sub-Byte Quantization}

The predominant datatype for NN inference on microcontrollers is 8-bit integer.
The use of smaller 4-bit datatypes~\cite{LSQuantization_ICLR2020,PROFIT_ECCV2020,thakker2021mlsys,annie1} for weights (activations) allows for more parameters (feature maps), potentially realizing higher accuracy in the same memory footprint.
However, current MCUs do not natively support sub-byte datatypes, so this must be emulated using $8$-bit types.
We investigated the benefit of 4-bit quantization on the KWS task.


Currently, the CMSIS-NN~\cite{lai2018cmsis}, does not provide convolution operators for $4-$bit values. 
Therefore, we developed optimized kernels for $4-$bit datatypes, and incorporated them into CMSIS-NN for use in our experiments.
This allows our DNAS to expand the search space to fit models with more weights and/or activations, potentially achieving higher accuracy in the same memory footprint.
The unpacking and packing routines required to emulate hardware support for $4-$bit using native $8-$ or $16-$bit operations add modest latency overhead.
These optimized kernels can efficiently support sub-byte quantization on either weights or activations or both. 
Prior work on mixed-precision inference (CMix-NN~\cite{capotondi2020cmix}) does not support operations on \textit{signed} sub-byte weight and activation values, nor non-modulo-$4$ feature-map channel numbers, and therefore is not compatible 
with current CMSIS-NN software and TFLM runtime stack.
We anticipate that future MCUs may provide native hardware support for $4-$bit datatypes, further increasing the value of this research direction.

\subsection{DNAS Backbones and Training Recipes}

DNAS requires a backbone supernet to be defined as the starting point for the search.
The design of the backbone is an important step which requires human experience of network operators and connectivity patterns that work well for a given task.
If the backbone is too large, the supernet will not fit in GPU memory.
On the other hand, if the backbone is too small, it may not provide a rich enough search space within which to find models that satisfy the constraints.
In this section, we describe the backbones used for the TinMLperf tasks and the training methodology.

\subsubsection{Visual Wake Words (VWW)}
We use a MobilenetV2 \cite{sandler2018mobilenetv2} backbone, consisting of a series of inverted bottleneck (IBN) blocks. Each IBN block includes the sequence: $1\times 1$ conv, $3 \times 3$ depthwise conv,  $1\times 1$ conv. We restrict our search space to the width of the first and last convolutions in each IBN, as well the convolutions preceding and following the sequence of IBN blocks. For each convolution, we choose between $10\%$ and $100\%$ of the width of the corresponding layer in MobilenetV2, in increments of $10\%$. In order to ensure that the input itself does not violate the SRAM constraint, we resize the input images to $50 \times 50 \times 1$ and $160 \times 160 \times 1$ for the small (STMF446RE) and medium (STM32F746ZG) sized MCUs, respectively. Note that we convert the RGB images to grayscale, such that the input only has $1$ channel, in order to trade off color resolution for spatial resolution \cite{fedorov2019sparse,chowdhery2019visual}. 

We run DNAS for 200 epochs, batch size 768, decaying the learning rate from $0.36$ to $0.0008$ with a cosine schedule. We use quantization aware training \cite{krishnamoorthi2018quantizing} to emulate $8-$bit quantization of both weights and activations during training. Discovered architectures are finetuned for 200 epochs with the same learning rate schedule, weight decay of $0.00004$, and knowledge distillation  using MobilenetV2 as the teacher, knowledge distillation coefficient $0.5$, and a temperature of $4$ \cite{hinton2015distilling}.

\begin{figure*}
    \centering
    
    \begin{subfigure}[b]{\textwidth}
         \centering
        \includegraphics[width=\textwidth]{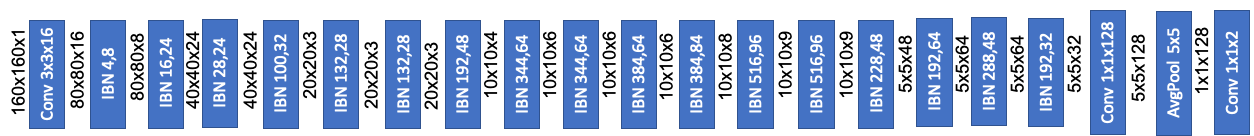}    
        \caption{}
         \label{fig:vww medium}
     \end{subfigure}
     
     \hfill
    \vspace{-10pt}
    
     \begin{subfigure}[b]{\textwidth}
         \centering
        \includegraphics[width=\textwidth]{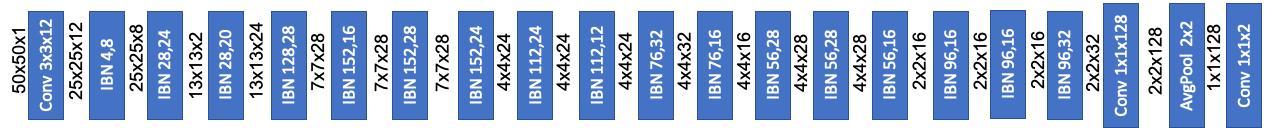}   
        \caption{}
         \label{fig:vww smallest}
     \end{subfigure}
     
     \vspace{-10pt}
     \caption{VWW architectures discovered by DNAS targeting (a) medium (STM32F446RE)  and (b) small (STM32F446RE) MCUs, labeled VWW-1 and VWW-2 in Appendix Table \ref{tab:full_results}, respectively. 
     The two numbers following IBN denote the number of expansion and compression filters.
     Tensor dimensions are provided in black text.
     }
    \label{fig:vww-archs}
\end{figure*}

\subsubsection{Keyword Spotting (KWS)}
After experimenting with different architectures on the KWS task, we settled on an enlarged DS-CNN(L) \cite{helloedge} model as the backbone for DNAS.
The backbone is built by adding four more depthwise-separable blocks of output channels 276 to the largest variant of DS-CNN. A skip connection branch (average pooling if the parallel convolutional block downsamples the input) is also added in parallel to each depthwise-separable block in the backbone to create shortcuts for choosing the number of layers. We use DNAS to choose the number of channels and the number of layers in this backbone network, while trying to satisfy the hardware constraints. 
The number of channels are restricted to multiples of 4 for good performance on hardware. 
For the small and medium models, the constrains were set to achieve 10FPS and 5FPS on the medium (STM32F746ZG) board while also suiting the smallest (STMF446RE) board. It is thus a combination of latency and working memory constraints. For the large model, we target latency of less than one second, in order to achieve real-time throughput.

DNAS is run for 100 epochs, with a batch size of 512, decaying the learning rate from $0.01$ to $0.00001$ with a cosine schedule. A weight decay coefficient of $0.001$ is used.  Additionally, we quantize weights, activations and input to 8-bit using fake quantization nodes to emulate deployment. The ranges of quantizers are learnt with gradient descent. 
We train the final models for another 100 epochs, with a batch size of $256$, decaying the learning rate from $0.02$ to $0.00008$ with a cosine schedule and a weight decay coefficient of $0.002$. SpecAugment~\cite{park2019specaugment} is used during training to further avoid overfitting.

\subsubsection{Anomaly Detection (AD)}
For AD, the model operates on spectrograms of audio signals in a similar way as for KWS, so it makes sense for the two tasks to share similar backbone networks. 
Hence, the backbone network we used for AD was DSCNN-L, with parallel skip connections (or average pooling if downsampling) to skip layers.
The strides of the last two depthwise-separable blocks are increased to 2 to downsample the input patch down to 4$\times$4 before applying the final pooling. DNAS searches for channel numbers and the total number of layers to meet the hardware deployment constraints. An anomaly detection system is expected to run in real-time for continuous monitoring, and 
should therefore take less time than the increment between two successive spectrogram images (considering overlapping). 
In our setting, this latency cutoff can be calculated as $32\times20\mathrm{ms}=640\mathrm{ms}$. This latency constraint together with the SRAM limits for each board are used as constraints in our DNAS runs. 

We use the same DNAS hyperparameters as for KWS, except we only train for 50 epochs, as convergence is faster. We also apply a mixup \cite{zhang2017mixup} augmentation coefficient of $0.3$ to avoid overfitting. We experimented with spectral warping augmentations~\citet{Giri2020}, but did not observe benefits in our setting. This may be 
because our models are relatively compact and use quantization aware training, and therefore require less data augmentation.

\vspace{-5pt}
\section{Results}

\subsection{Methodology}
To deploy our models, we convert them to TFlite format and then execute them on each MCU using the TFLM runtime. 
The eFlash occupancy is determined using the Mbed compiler~\cite{mbed-os} and the SRAM consumption is obtained using the TFLM recording memory APIs. We measure latency on the MCU using the Mbed Timer API. 

\subsection{Visual Wake Words (VWW)}

\begin{figure}[t]
    \includegraphics[width=\linewidth]{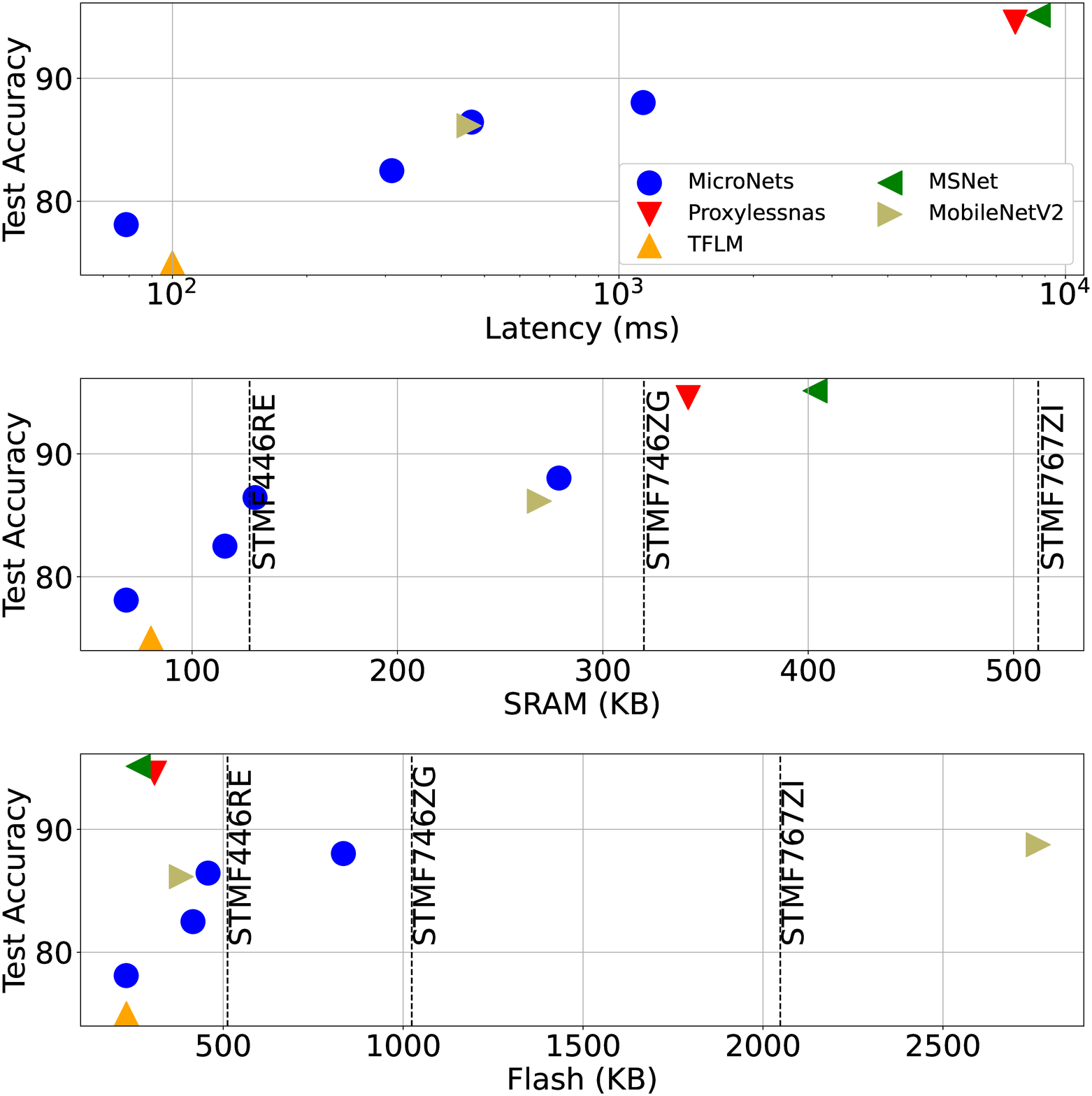}
\vspace{-25pt}
    \caption{
    VWW results targeting the small and medium MCUs. Our search yields pareto optimal MicroNets.
    Latency measured on the STM32F746ZG. SRAM and Flash refer to the overall measured usage of the model, without the TFLM overheads Figure \ref{fig:mem-map}.
    }
    \label{fig:VWW}

\end{figure}

Figure~\ref{fig:VWW} compares our DNAS results (MicroNets) to three state of the art results, including ProxylessNAS \cite{cai2018proxylessnas}, MSNet \cite{cheng2019msnet}, and the TFLM example model \cite{chowdhery2019visual}. The largest network in our search space is MobileNetV2, which achieves $88.75\%$ accuracy. The MicroNet models are visualized in Figure \ref{fig:vww-archs}. We found that the model produced by targeting the medium MCU ($88.03\%$) nearly matched the accuracy of MobileNetV2, obviating the need to search for a large-MCU specific model. MicroNets are pareto-optimal for the small and medium sized MCUs. For the small MCU, our MicroNet is $3.1\%$ more accurate than the TFLM reference, the only network considered which can be deployed on the small MCU with TFLM, while being $21$ms faster. For the medium MCU, our MicroNet model was the only model considered that could be deployed on that MCU. 

\begin{figure}[t]
    \includegraphics[width=\linewidth]{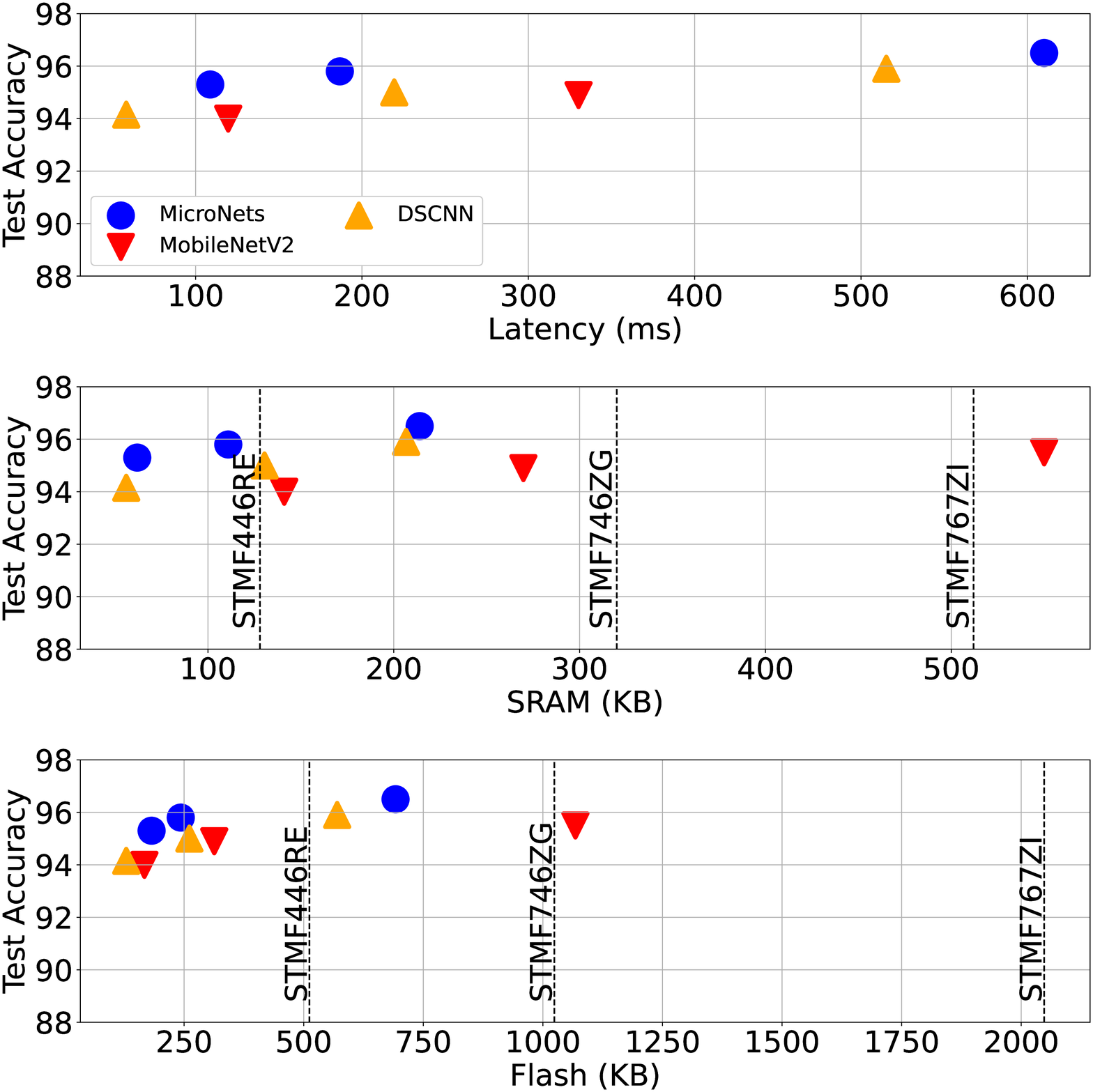}
\vspace{-25pt}
\caption{
KWS results. The small and medium MicroNets both target the smallest MCU.
MicroNet medium model is more accurate than DS-CNN(L) and is $2.7\times$ faster. 
The largest MobileNetV2 variant does not fit and is omitted.
Latency measured on the STM32F746ZG. SRAM and Flash refer to the overall measured usage of the model, without the TFLM overheads (Figure~\ref{fig:mem-map}). 
}
    \label{fig:kws}

\end{figure}

The limitations of previous TinyML models is clear in Figure \ref{fig:VWW}: they do not fully exploit the precious memory resources.
For instance, the ProxylessNAS model easily fits the flash memory on the smallest MCU (STM32F446RE), but requires the largest MCU (STM32F767ZI) to fit the activations in SRAM.
Therefore, ProxylessNAS will only run on the large MCU.
MSNet shows similar characteristics.
These limitations underline the motivation for DNAS optimized models to target a specific MCU size.

MicroNet models are not tight against the SRAM and flash constraints for a number of reasons.
TFLM has to schedule every graph and perform memory management, which leads to some variability in the resulting model size.  
Therefore, we estimate the maximum model size and activation footprint possible for a given hardware platform by performing some experiments with TFLM.
However, the final binary size is still somewhat dependent on the graph itself, so this prevents us from tightly meeting the constraints.
In a real application context there will also be application logic and potentially even a real-time operating system (RTOS)~\cite{mbed-os}, which will take additional eFlash memory resources that must be budgeted into the model constraints.

\vspace{-5pt}
\subsection{Keyword Spotting (KWS)}
\label{sec:KWS_results}
KWS results are shown in Figure~\ref{fig:kws}, where we compare our results (MicroNet) to DS-CNNs \cite{helloedge} and models built by stacking MobileNetV2 \cite{sandler2018mobilenetv2} inverted bottleneck blocks. Few results are currently available for version 2 of the Google speech commands dataset, therefore we train baseline models for comparison. We trained all the models with exactly the same training recipe and quantized them to $8-$bit weights and activations (including input) before measuring their accuracy.  MicroNet models are the Pareto optimal for latency, SRAM usage and model size. 
MicroNet small and medium models also come very close to the latency constraints set for them, achieving 9.2FPS and 5.4FPS on the medium sized MCU while having accuracy of $95.3\%$ and $95.8\%$ and being deployable on the smallest MCU. Detailed description of MicroNet KWS model architectures can be found in Table \ref{tab:architectures} of Appendix \ref{Appendix:A}. 

We can further leverage sub-byte quantization to make bigger but more accurate models deployable on smaller MCUs.
Table~\ref{table:4b_micronets} demonstrates the accuracy, latency, and SRAM memory trade-off of the $4-$bit MicroNet-KWS-Large model. 
The $4-$bit MicroNet-KWS-Large model outperforms the $8-$bit medium-sized model by $0.5\%$, because it is able to have more weights and activations on the same MCU. Table~\ref{table:4b_micronets} reports the model latency as measured on the medium MCU (STM32F746ZG).
The increase in latency of the $4$-bit model is primarily attributed to increase in ops due to larger feature maps.
When compared to the sub-byte kernels of CMix-NN~\cite{capotondi2020cmix}, our $4-$bit kernels can substantially hide the latency overhead due to software-emulation of $4-$bit operations, by fully exploiting the available instruction-level-parallelism bandwidth on Cortex-M microcontrollers. 
Furthermore, we believe the accuracy of the $4-$bit KWS MicroNet can be further improved by selectively quantizing lightweight depthwise layers to $8-$bits, while quantizing remaining memory- and latency-heavy pointwise and standard convolutional layers to $4$-bits~\cite{RusciMixedPrecision2020,GopeCVPR2020}.
Latency results for 4-bit quantized MicroNet-KWS models and their comparison against $8$-bit models can be found in the Appendix.

\subsection{Anomaly Detection (AD)}
\label{sec:AD_results}
Results for AD are given in Table~\ref{table:AD_results}. To aid comparison, we use the ``Uptime" metric defined as model latency divided by stride time between two successive inputs. 
For real-time AD, this ratio is the duty cycle of the MCU workload. 
For example, our models expect a stride of $640\mathrm{ms}$ while some other models expect $32\mathrm{ms}$ and need to be run more often. This metric is also relates directly to power consumption.

\begin{table}[t]
\centering
\caption{
KWS results for 4-bit quantized MicroNet models.
}
\scriptsize
\begin{tabular}{l| c | c | c }
\toprule
& MN-KWS-L & MN-KWS-M & MN-KWS-L \\
& (8-b W/8-b A) & (8-b W/8-b A) & (4-b W/4-b A) \\
\midrule
Accuracy (\%) & 96.5 & 95.8 & 96.3 \\
Latency (s) (M) & 0.59 & 0.18 & 0.76 \\
Model Size (KB) & 612 & 163 & 375* \\
SRAM (KB) & 208 & 103 & 121* \\
\bottomrule
\end{tabular}
\\
* denotes estimated value.
\vspace{-15pt}
\label{table:4b_micronets}
\end{table}

The models obtained from DNAS are called MicroNet-AD. The three different sized models each targets a different sized MCU. Our solutions are compared with a baseline fully-connected auto-encoder (FC-AE) for anomaly detection \cite{purohit2019mimii}, which has a 640 dimensional input, followed by 4 fully-connected hidden layers of 128 neurons each, a bottleneck layer of 8 neurons, 4 fully-connected hidden layers of 128 neurons again and the output. This baseline model achieves 84.76\% AUC and runs fast. However, once we try to scale it up for better anomaly detection performance, the size of the model quickly exceeds the flash limit of all MCUs used in this work. The wide FC-AE model which scales up all the hidden neuron number from 128 to 512 in the baseline, achieves only 87.1\% AUC but its size exceeds 2MB in 8-bit, making it undeployable on our MCUs. An alternative to fully-connected AE that is more parameter efficient is convolutional AE, which we have also included in the comparison. 
Convolutional AEs require the transposed convolution operator, not supported in TFLM.

We also compare with the MobileNetV2-0.5AD model trained in a similar self-supervised way as ours, which is a component of the winning solution at DCASE2020 challenge \cite{DCASE2020} presented in \citet{Giri2020}. Since the authors submitted ensembles of multiple classifiers, we take the average of AUCs reported where the MobileNetV2-0.5AD is a component as an estimate of its accuracy. This model can only be deployed on the largest MCU because of it relatively large size (close to 1MB). The model is light in number of operations but its uptime requirement is worse than our solutions since it expects a time stride of $256\mathrm{ms}$. Our large MicroNet model is equally performing in terms of AUC, requires less than half the Flash size and consumes less compute resources in terms of uptime requirement. The smallest MicroNet-AD model can be deployed on the small MCU performing real-time AD with $>95\%$ AUC performance. As we have shown previously, the small MCU only draws about $1/3$ the power of the medium, which is attractive since most of these tiny IoT devices run on batteries or need to be energy self-sufficient. MicroNet-AD model architectures can be found in Table \ref{tab:architectures} of Appendix \ref{Appendix:A}.
\vspace{-5pt}

\begin{table}[t]
\centering
\caption{
AD results. 
AUC for MicroNets is with 8-bit weights/activations; Conv-AE~\cite{ribeiro2020deep} and MBNETV2-0.5AD~\cite{Giri2020} are unquantized FP32.
S, M and L denote small, medium and large MCU targets.
}
\scriptsize
\resizebox{\columnwidth}{!}{
\begin{tabular}{l| c | c | c | c| c }
\toprule
Models & AUC(\%) & Ops(M) & Size & Mem & Uptime($\%$) \\
\midrule
MicroNet-AD(L) & 97.28 & 129 & 442KB & 383KB & 95.9\,(L)\\
MicroNet-AD(M) & 96.22 & 124.7 & 464KB & 274KB & 94.8\,(M)\\
MicroNet-AD(S) & 95.35 & 37.5 & 253KB & 114KB & 71.4\,(S)\\
FC-AE(Baseline) & 84.76 & 0.52 & 270KB & 4.7KB & 10.3\,(M)\\
FC-AE(Wide) & 87.1 & 4.47 & 2.2MB & \;\;4.7KB* & ND\\
Conv-AE & 91.77 & 578 & \;\;4.1MB* & \;\;160KB* & ND\\
MBNETV2-0.5AD & \;\;97.24* & 31.1 & 965KB & 206KB & 98.8\,(L)\\
\bottomrule
\end{tabular}}
* denotes estimated value.  ND denotes not deployable on MCU. 
\label{table:AD_results}
\end{table}

\subsection{Comparison with State-of-the-Art}
A key previous TinyML work is SpArSe~\cite{fedorov2019sparse}.
This work is focused on even smaller MCUs, with memory down to 2KBs.
However, it also targets smaller datasets with smaller input dimensions than the tinyMLperf tasks that we use in our work.
In a parallel line of work, MCUNet \cite{lin2020mcunet} demonstrated SOTA MCU models using a framework that jointly designs the model architecture and the lightweight code-generation inference engine. 
Their latency and SRAM measurement relies on a closed-source software stack that is not available to us, so it is difficult to make comparison with their results.
However, our models are pareto optimal compared to MCUNet on the KWS task even with a readily available, open source software stack (see Figure \ref{fig:MCUNet_KWS} in Appendix \ref{Appendix:MCUNet_comp}).

In support of this paper, we have open sourced our models at \url{https://github.com/ARM-software/ML-zoo}.
We hope these will be useful for MCU vendors and researchers, as a set of standard models for benchmarking.
\vspace{-10pt}

\section{Conclusion}

TinyML promises to enable a broad array of IoT applications, but is technically challenging.
This is primarily due to the memory demands of deep neural network inference, which are at conflict with the limitations of MCUs.
We start by analyzing measured MCU inference performance.
Measurements demonstrate that for models sampled from a given network search space, the inference latency of the model is, in fact, linear with the total operation count.
Since MCU power is largely independent of workload, operation count is also a strong proxy for energy per inference.
Therefore, we use operation count as a proxy for both latency and energy, and setup a differentiable NAS search to design a family of models called MicroNets.
MicroNet models optimized for multiple MCUs demonstrate state-of-the-art performance on all three tinyMLperf tasks: visual wake words, audio keyword spotting and anomaly detection.

\section*{Acknowledgements}
This work was sponsored in part by the ADA (Applications Driving Architectures) Center.




\bibliography{MicroNets}
\bibliographystyle{mlsys2021}


\appendix

\begin{table*}[t]
  \caption{Results Table. (*) Estimated (-) Unable to measure do to SRAM or eFlash constraints. ProxylessNas ~\cite{cai2018proxylessnas}, MSNet~\cite{cheng2019msnet}, Person Detection~\cite{TFLM}, and MBNetV2-0.5~\cite{Giri2020} are all previous work.}
  \label{tab:full_results}
  \centering
  \scriptsize
\resizebox{\textwidth}{!}{
\begin{tabular}{l|c|c|c|c|c|c|c|c|c|c|c}\toprule
\textbf{Dataset} &\textbf{Model} &\textbf{Accuracy} &\textbf{Binary} &\textbf{Flash} &\textbf{SRAM} &\textbf{Mops} &\textbf{S Latency} &\textbf{M Latency} &\textbf{L Latency} &\textbf{S Energy} &\textbf{M Energy} \\
 & & &\textbf{(KB)} &\textbf{(KB)} &\textbf{(KB)} &\textbf{Mops} &\textbf{(Sec)} &\textbf{(Sec)} &\textbf{(Sec)} &\textbf{(mJ)} &\textbf{(mJ)} \\
\midrule
GSC &MicroNet-KWS-L &96.5 &701 &612 &208.8 & 129 &- &0.610 &0.596 &- &274.32 \\
GSC &MicroNet-KWS-M &95.8 &252 &163 &103.3 &30.6 &0.426 &0.187 &0.181 &70.56 &83.16 \\
GSC &MicroNet-KWS-S &95.3 &191 &102 &53.2 &16.4 &0.250 &0.109 &0.108 &40.68 &48.6 \\

VWW & MicroNet-VWW-1 & 88.03 &949  & 833 & 285.3 & 135.9 & - & 1.133 &1.055  &-  &478.8 \\
VWW & MicroNet-VWW-2 & 78.1 &331  & 230 & 69.5 & 5.3 & 0.181 & 0.079 &0.082 &27.25  &36.36 \\
VWW & MicroNet-VWW-3 & 86.44 &564  & 458 & 133.7 & 45.2 & - & 0.467 &0.447 &-  &196.2 \\
VWW & MicroNet-VWW-4 & 82.49 &521  & 416 & 118.7 & 37.7 & 0.726 & 0.31 &0.298 &-  &133.2 \\

MIMII &MicroNet-AD-L &AUC: 97.28 &530 &442 &383.7 &129 &- &- &0.614 &- &- \\
MIMII &MicroNet-AD-M &AUC: 96.05 &562 &464 &274.5 &124.7 &- &0.608 &0.567 &- &269.64 \\
MIMII &MicroNet-AD-S &AUC: 95.35 &351 &253 &114.2 &37.5 &0.457 &0.192* &0.194 &74.16 &91.8 \\
\hline
GSC &DSCNN-L &95.9 &579 &490 &201.3 &107.2 &- &0.515 &0.497 &- &229.32 \\
GSC &DSCNN-M &95 &270 &181 &123.3 &37.3 &- &0.219 &0.212 &- &98.64 \\
GSC &DSCNN-S &94.15 &138 &49 &47.2 &7.1 &0.131 &0.058 &0.058 &21.132 &25.956 \\
GSC &MBNETV2-L &95.5 &- &988 &530* &276.8 &- &- &- &- &- \\
GSC &MBNETV2-M &94.9 &331 &233 &266.0 &59.26 &- &0.330 &0.317 &- &147.6 \\
GSC &MBNETV2-S &94 &185 &87 &134.2 &16.1 &- &0.120 &0.115 &54 &15.264 \\
VWW &ProxylessNas &94.6 &413 &309 &349.8 &- &- &7.72* &7.543 &- &- \\
VWW &MSNet &95.13 &362 &264 &413.0 &- &- &8.69* &8.499 &- &- \\
VWW &Person Detection &76 &398 &294 &82.3 &- &0.254 &0.108 &0.108 &39.96 &49.32 \\
MIMII &AD-baseline &AUC:84.76 &346 &270 &4.7 &0.52 &0.007 &0.003 &0.003 &1.1736 &1.26 \\
MIMII &MBNetV2-0.5 &AUC: 97.24* &1050 &965 &206.8 &31.1 &- &- &0.253 &- &- \\
\bottomrule
\end{tabular}}
\end{table*}

\begin{table*}[t]
  \centering
\caption{Model architectures for keyword spotting and anomaly detection MicroNet models, $\mathrm{h}$ and w are height and width of the convolutional filters, c is the number of output channels and s is the stride.}\label{tab:architectures}
  \scriptsize
\resizebox{\textwidth}{!}{
\begin{tabular}{l|c|c}\toprule
\textbf{Dataset} &\textbf{Model} &\textbf{Architecture}
\\
\midrule
GSC &MicroNet-KWS-L &
\shortstack{Conv2D (h:10,w:4,c:276,s:1)-Depthwise Separable Block (h:3,w:3,c:248,s:2)-Depthwise Separable Block (h:3,w:3,c:276,s:1)\\
-Depthwise Separable Block (h:3,w:3,c:276,s:1)-Depthwise Separable Block (h:3,w:3,c:248,s:1)-
Depthwise Separable Block (h:3,w:3,c:248,s:1)\\
-Depthwise Separable Block (h:3,w:3,c:248,s:1)-Depthwise Separable Block (h:3,w:3,c:248,s:1)-AvgPool(h:25, w:5,s:1)-FC(c:12)} \\
\midrule
GSC & MicroNet-KWS-M
&
\shortstack{Conv2D (h:10,w:4,c:140,s:1)-Depthwise Separable Block (h:3,w:3,c:140,s:2)-Depthwise Separable Block (h:3,w:3,c:140,s:1)\\
-Depthwise Separable Block (h:3,w:3,c:140,s:1)-Depthwise Separable Block (h:3,w:3,c:112,s:1)-
Depthwise Separable Block (h:3,w:3,c:196,s:1)\\
-AvgPool(h:25, w:5,s:1)-FC(c:12)} \\
\midrule
GSC & MicroNet-KWS-S
&
\shortstack{Conv2D (h:10,w:4,c:84,s:1)-Depthwise Separable Block (h:3,w:3,c:112,s:2)-Depthwise Separable Block (h:3,w:3,c:84,s:1)\\
-Depthwise Separable Block (h:3,w:3,c:84,s:1)-Depthwise Separable Block (h:3,w:3,c:84,s:1)-
Depthwise Separable Block (h:3,w:3,c:196,s:1)\\
-AvgPool(h:25, w:5,s:1)-FC(c:12)} \\
\midrule
MIMII & MicroNet-AD-L
&
\shortstack{Conv2D (h:3,w:3,c:276,s:1)-Depthwise Separable Block (h:3,w:3,c:248,s:2)-Depthwise Separable Block (h:3,w:3,c:276,s:1)\\
-Depthwise Separable Block (h:3,w:3,c:276,s:1)-Depthwise Separable Block (h:3,w:3,c:248,s:2)-
Depthwise Separable Block (h:3,w:3,c:248,s:2)\\
-AvgPool(h:4, w:4,s:1)-FC(c:4)}\\
\midrule
MIMII & MicroNet-AD-M
&
\shortstack{Conv2D (h:3,w:3,c:192,s:1)-Depthwise Separable Block (h:3,w:3,c:276,s:2)-Depthwise Separable Block (h:3,w:3,c:276,s:1)\\
-Depthwise Separable Block (h:3,w:3,c:276,s:1)-Depthwise Separable Block (h:3,w:3,c:276,s:2)-
Depthwise Separable Block (h:3,w:3,c:276,s:2)\\
-AvgPool(h:4, w:4,s:1)-FC(c:4)}\\
\midrule
MIMII & MicroNet-AD-S
&
\shortstack{Conv2D (h:3,w:3,c:72,s:1)-Depthwise Separable Block (h:3,w:3,c:164,s:2)-Depthwise Separable Block (h:3,w:3,c:220,s:1)\\
-Depthwise Separable Block (h:3,w:3,c:276,s:2)-Depthwise Separable Block (h:3,w:3,c:276,s:2)\\
-AvgPool(h:4, w:4,s:1)-FC(c:4)}\\
\bottomrule
\end{tabular}
}
\end{table*}


\section{Model architectures for KWS and AD models} \label{Appendix:A}
In Table \ref{tab:architectures}, we report the detailed architectures of MicroNet models for keyword spotting and anomaly detection found through DNAS.

\section{Results Table}
In Table \ref{tab:full_results} we provide a table of our results and baselines for easy comparison with future work.
Binary refers to the size of the compiled binary that is loaded onto the MCU.
Flash is the flash consumption of the model and SRAM is the total SRAM consumption of the model.
We also report latency on the STM32F446RE (S), STM32F746ZG (M) and SRM32F767ZI (L) as well as energy consumption on the STM32F446RE (S) and STM32F746ZG (M).
All of models are deployed using the TFLM inference framework. 
The Flash consumption is determined by the size of the tflite flatbuffer file and the SRAM consumption is obtained using the TFLM recording micro interpreter.
We measure latency on the MCU in microseconds using the MBED Timer API. 
Finally we use the Qoitech Otii Arc~\cite{otii} to measure energy consumption.

\section{Power Trace}
\begin{figure}[t]
    \includegraphics[width=\linewidth]{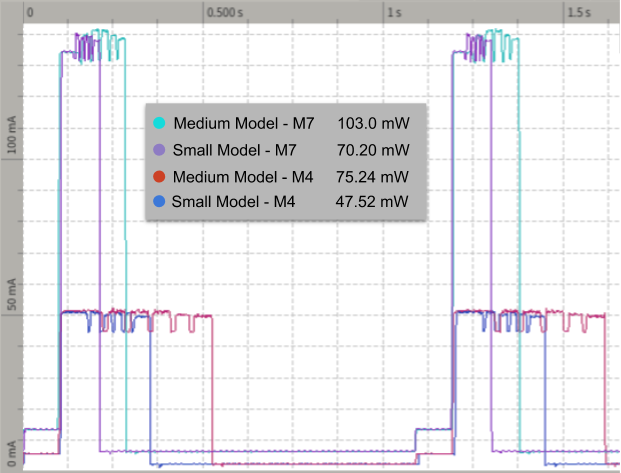}
     \vspace{-20pt}
     \caption{
     Current consumption of a small and medium model on the STM32f446RE and STMF746ZG. We report the average power consumption measured over one second, including active and idle power.
     }
     \label{fig:model_power}
\end{figure}

We plot the current vs. time for a small model and a medium model on the STM32f446RE and STMF746ZG in Figure \ref{fig:model_power}.
We also report the average power consumption over 1 second to illustrate the impact of the deep sleep power consumption on the overall energy consumption of a tinyML application with a duty cycle of one frame per second.
We show that the current consumption varies little between models but the smaller model consume significantly less energy due to its reduced latency.
Figure \ref{fig:model_power} also demonstrates that the smaller mcu consumes less power on average despite being active for longer.

\section{Latency Measurement for 4-bit MicroNet Models}

\begin{figure}[t]
    \includegraphics[width=\linewidth]{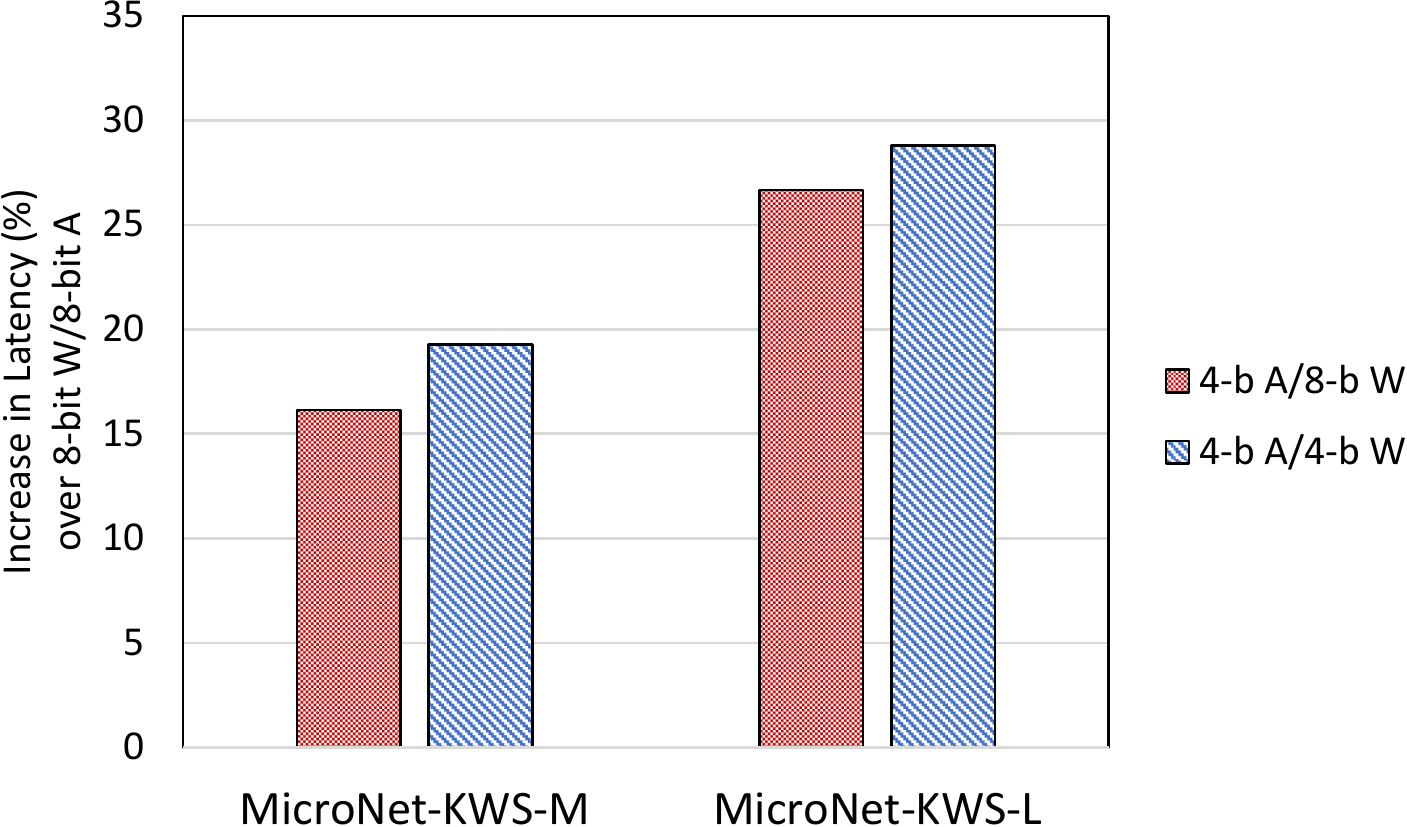}
     \caption{Percentage increase in latency of the different models with $4$-bit quantization on the medium MCU (STM32F746ZG) board in comparison to their $8$-bit counterparts.
     }
     \label{fig:4bit_vs_8bit_micronets_latency}
\end{figure}
Figure~\ref{fig:4bit_vs_8bit_micronets_latency} reports the percentage increase in latency of the MicroNet-KWS-M and MicroNet-KWS-L models as $4$-bit quantization is applied to either weights or activations or both. $4$-bit A/$4$-bit W refers to our optimized kernels that emulate $4$-bit support on both weights and activations, while $4$-bit A/$8$-bit W denotes kernels that emulate $4$-bit datatypes support only on activations. It is important to note that the latency increase using our $4$-bit optimized kernels is marginal even for MicroNet-KWS-M and MicroNet-KWS-L like deep networks ($19.28\%$ and $28.8\%$ increase for MicroNet-KWS-M and MicroNet-KWS-L respectively over their $8$-bit quantized models). Any smaller network than these (e.g. MicroNet-KWS-S, etc.) should observe even lower increase in latency with our $4$-bit quantized kernels.

\section{Comparison with MCUNet on KWS}\label{Appendix:MCUNet_comp}
\begin{figure}[t]
\centering
    \includegraphics[width=.72\linewidth]{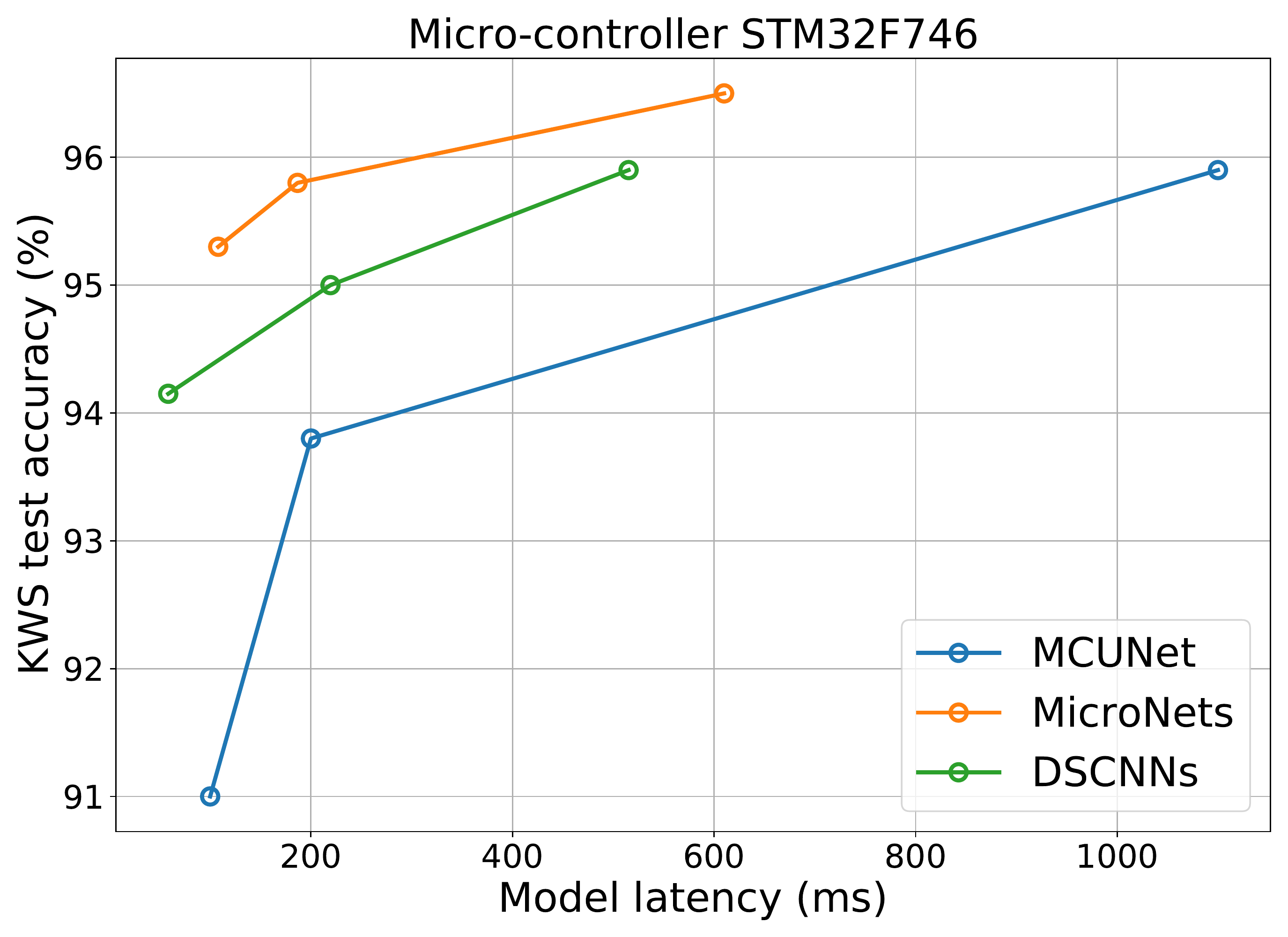}
    \includegraphics[width=.72\linewidth]{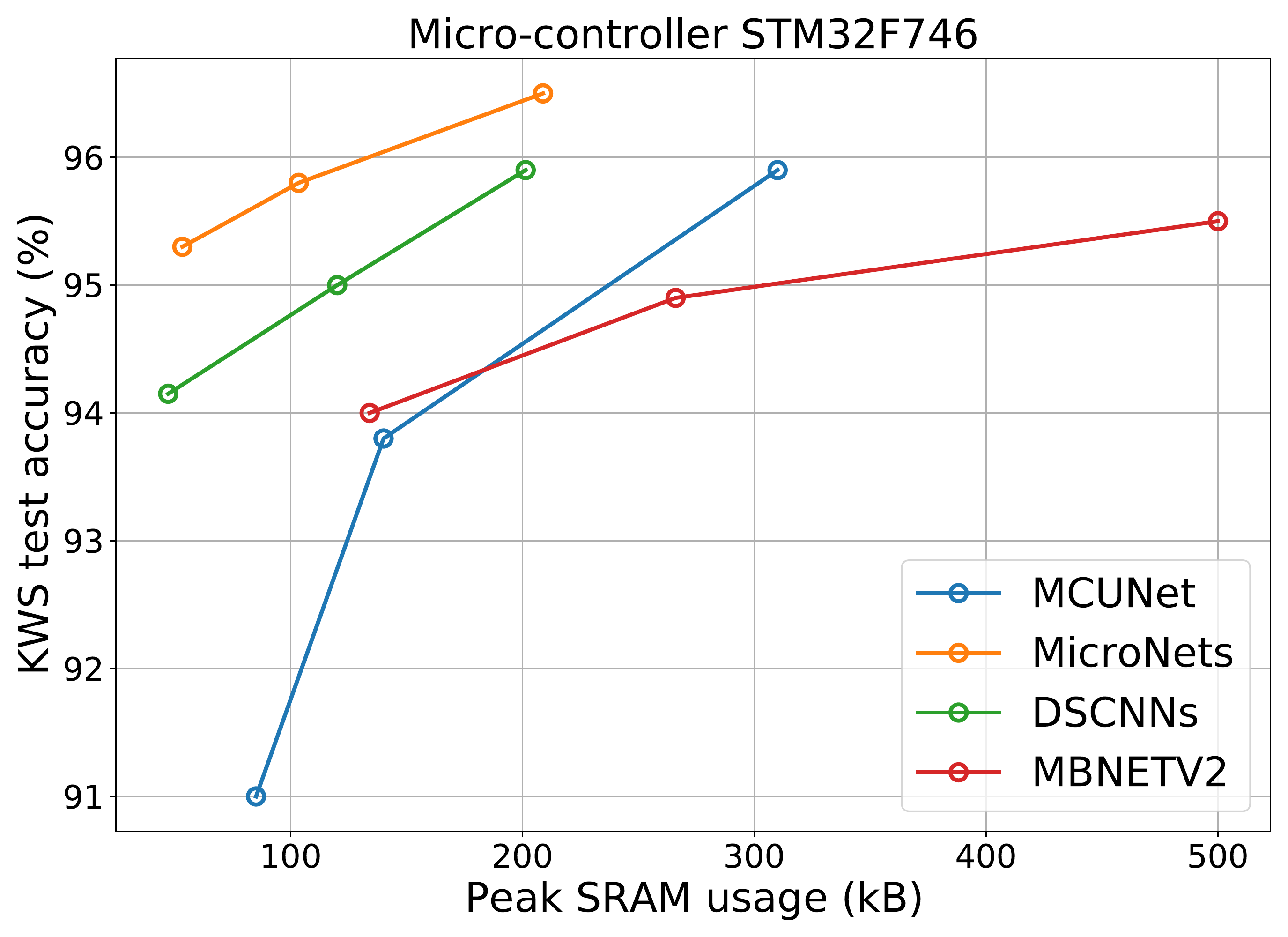}
    \caption{Comparison of different KWS models running on the medium MCU (STM32F746ZG).}
    \label{fig:MCUNet_KWS}
\end{figure}

A Pareto front comparison between MicroNets and MCUNet models on the KWS task is shown in Figure \ref{fig:MCUNet_KWS}, the data points for the MCUNet KWS models are our best estimates from figures published in \citet{lin2020mcunet}.

\end{document}